\pdfoutput=1                

\documentclass[12pt,letterpaper]{article}

\usepackage[utf8]{inputenc}
\usepackage[T1]{fontenc}
\usepackage{lmodern}            
\usepackage[affil-it]{authblk} 
\usepackage{enumitem}
\usepackage{float}
\usepackage{amsmath}          
\DeclareMathOperator{\logit}{logit}
\usepackage{geometry}
\usepackage{setspace}
\usepackage{amsmath,amssymb,amsfonts,amsthm}

\usepackage{graphicx}
\usepackage[table,dvipsnames]{xcolor}

\usepackage{booktabs,longtable,multirow,array}

\newcolumntype{C}[1]{>{\centering\arraybackslash}m{#1}}

\usepackage[colorlinks=true,
            linkcolor=BrickRed,
            citecolor=SeaGreen,
            urlcolor=MidnightBlue]{hyperref}

\usepackage[font=footnotesize,labelfont=bf,labelsep=period]{caption}

\usepackage[numbers,sort&compress]{natbib}
\definecolor{bloodorange}{HTML}{D15500}
\definecolor{darkgreen}{HTML}{004d00}

\newcommand{\largesc}[1]{{\fontsize{12pt}{14pt}\selectfont
  \textbf{\textsc{\textcolor{bloodorange}{#1}}}}}

\title{\textbf{\textcolor{bloodorange}{Real‑World Gaps in AI Governance Research}}\\[0.25em]\large AI safety and reliability in everyday deployments}
\author[1,2]{Ilan Strauss}
\author[1]{Isobel Moure}
\author[1,3]{Tim O'Reilly}
\author[1]{Sruly Rosenblat\thanks{We gratefully acknowledge funding support from The Alfred P. Sloan Foundation, the Omidyar Network, and the Patrick J. McGovern Foundation.\\Contact: \href{mailto:istrauss@ssrc.org}{istrauss@ssrc.org}.\\Code and data: \url{https://github.com/AI-Disclosures-Project/The-State-of-AI-Governance-Research}}}
\affil[1]{AI Disclosures Project, Social Science Research Council}
\affil[2]{Institute for Innovation and Public Purpose, University College London}
\affil[3]{O'Reilly Media}
\date{}

\begin{document}

\maketitle

\begin{abstract}\onehalfspacing
Drawing on 1,178 safety and reliability papers from 9,439 generative AI papers (January 2020 – March 2025), we compare research outputs of leading \textit{AI companies} (Anthropic, Google DeepMind, Meta, Microsoft, and OpenAI) and \textit{AI universities} (CMU, MIT, NYU, Stanford, UC Berkeley, and University of Washington). We find that corporate AI research increasingly concentrates on pre‑deployment areas—model alignment and testing \& evaluation—while attention to deployment‑stage issues such as model bias has waned. Significant research gaps exist in high‑risk deployment domains, including healthcare, finance, misinformation, persuasive and addictive features, hallucinations, and copyright. Without improved observability into deployed AI, growing corporate concentration could deepen knowledge deficits. We recommend expanding external researcher access to deployment data and systematic observability of in‑market AI behaviors.

\end{abstract}

\thispagestyle{empty}
\newpage

\tableofcontents
\thispagestyle{empty}
\newpage
\setcounter{page}{1}
\pagenumbering{arabic}

\section{Introduction}
\setcounter{page}{1}
\pagenumbering{arabic}

As generative AI becomes integrated into every facet of our work and social lives, there is an \textcolor{bloodorange}{urgent need to understand the performance and impact of AI products in such commercial ``post-deployment'' contexts} \citep{chayes2025draft, weidinger2025toward}. Yet corporate research, now increasingly dominant, focuses on AI risks in \textit{pre-deployment} laboratory settings through model alignment and testing (Figure \ref{fig:cite_v_sample}).\footnote{AI alignment covers `post-training' interventions, fine-tuning \& reinforcement learning from human and AI feedback.} User, system, and society-level impacts remain neglected.\footnote{AI companies' do revise their models based based on red-teaming and user experience feedback \citep{openai2024o1systemcard}.}

\textcolor{bloodorange}{Unless AI governance research follows AI systems into the real world, areas currently considered highest risk by AI companies themselves will remain underexplored}. These include model persuasiveness, emergent behaviors from reinforcement learning exploitation (`reward hacking'), and misinformation \citep{phuong2024evaluating, weidinger2024holistic, jaech2024openai}. De-prioritization of \textit{research} into such areas both impedes developing industry-wide best \textit{practices} for deployed AI systems and confines essential AI safeguards to siloed corporate efforts, limiting knowledge diffusion and public accountability.

\textcolor{bloodorange}{Growing corporate concentration in AI research risks exacerbating these deficiencies}. The commercial `AI race' prioritizes an engaging user experience over broader societal impacts \citep{WSJ2025MetaChatbots}. Evidence of this shift includes corporate research teams becoming tightly integrated with product teams \citep{techcrunch2025google}, research findings increasingly kept internal \citep{heikkila2025deepmind} (Figure \ref{fig:time_Series}), and alignment research overlooking dangerous side-effects, such as sycophancy and degraded answer quality \citep{amodei2016faulty, sycophancy2023, denison2024sycophancy, zeff2025openai}.\\[-4mm] 

\noindent \largesc{Method}\\[-7mm]

\textcolor{bloodorange}{\textit{We analyze AI governance research using a dataset of 1,178 safety and reliability papers from 9,439 generative AI papers}} written by five dominant \textit{AI companies} (Anthropic, Google DeepMind, Meta, Microsoft, and OpenAI), and six prominent \textit{AI research universities} (Carnegie Mellon University (CMU), MIT, New York University (NYU), Stanford, UC Berkeley, and University of Washington) between January 2020 and March 2025. We call these two groups `Corporate AI' and `Academic AI', respectively. Our dataset combines generative AI research papers from Anthropic and OpenAI's websites \citep{delaney2024mapping} with OpenAlex's database. We define AI governance research as technical and applied safety and reliability research pre- and post-deployment. In conjunction with OpenAI's o3-mini, we determine if papers are ``safety \& reliability'' research, and then classify them into one of eight sub-categories. We also conduct separate `regex' key word searches in paper abstracts and titles for high-risk deployment domains (medical, finance, commercial, \& copyright) and capabilities (misinformation, disclosures, behavioral, \& accuracy).\\[-3mm] 

\noindent\largesc{Core Findings}
\vspace{-1.5mm}
\begin{enumerate}[itemsep=4.4pt, parsep=0.2pt]

\item \textcolor{bloodorange}{\textbf{\textit{AI governance research is highly concentrated within a handful of uniquely resourced and integrated AI tech companies, with a disproportionately influential research impact}}}. Anthropic, OpenAI, and Google DeepMind each have far more citations for their AI safety \& reliability work than any of the major U.S. academic institutions we track. Google DeepMind has more citations for its general generative AI research than the top four AI academic institutions combined.

\item \textcolor{bloodorange}{\textbf{\textit{As leading AI companies race to commercialize powerful AI systems, their research priorities are increasingly shaped by business incentives rather than by comprehensive risk assessments and mitigations}}}. Most of the corporate governance research we review focuses on model performance divorced from its applications. Ethics \& bias research -- needed to understand systematic, unjustified differences in LLM behavior or outputs -- now only receives attention from academic researchers. 

\item \textcolor{bloodorange}{\textbf{\textit{Corporate AI labs severely neglect deployment-stage behavioral and business risks}}}. Only 4\% of Corporate AI papers (6\% Academic AI) tackle high-stakes areas like persuasion, misinformation, medical \& financial contexts, disclosures, or core business liabilities (IP violations, coding errors, hallucinations) -- despite emerging lawsuits showing these risks to already be material.

\end{enumerate}

\noindent \largesc{Policy Considerations} 

\textcolor{bloodorange}{\textbf{\textit{To guard against commercialization-driven risks, third-party researchers (and auditors) need data on AI systems operating in real-world environments}}}. Commercial incentives drive innovation but also foster corporate risk-taking, potentially lowering safeguards when they conflict with profit-maximizing business models \citep{WSJ2025MetaChatbots, Edwards2025AI}. \textit{Post-deployment monitoring research is therefore publicly vital but currently limited to piecemeal AI incident databases} \citep{marchal2024generative, willison2024owasp, MIT2024AIIncidentTracker}, \textit{old or overly aggregated user-LLM chat data} \citep{anthropic_clio_2024, zhao2024wildchat}, and public testing of models. Real-world visibility into the effects of AI systems is negligible.

\textcolor{bloodorange}{\textbf{\textit{Structured access is needed into deployed AI systems' telemetry data and artifacts to systematically analyze real-world risks and harms.}}}
Monitoring and evaluation of LLMs in real-world environments is now essential to quality assurance (QA), as in `LLMOps' \citep{aryan2024llmops}. But the data used for this is the preserve of corporate \textit{practice}, resulting in society losing essential insight into AI's ongoing risks and harms. Disclosure of AI system \textit{telemetry data} (logs, traces, \& business metrics) and LLM model \textit{data artifacts} (e.g., training/fine-tuning datasets) may expose corporations to liability. But emerging LLM monitoring frameworks -- such as those from LangSmith, Langfuse, OpenTelemetry, \& Weights and Biases -- make structured \& standardized external API access for researchers increasingly feasible. Liability safe harbors \citep{longpre2024safe, arcila2025ai} are likely required to support purpose-built external access; otherwise, deployment research will have to rely on public-private partnerships.\\[-4mm]

\noindent \textbf{Literature and Roadmap}. Important papers in AI research classification are \citet{toner2022exploring}, \citet{2023impactsresearch}, \citet{cottier2023leading}, \citet{klyman2024expanding} -- and most recently \citet{delaney2024mapping}, which addresses pre-deployment technical AI safety research only. Next, Section \ref{sec:methods} motivates our study's focus on AI's deployment, and describes our data and method (Appendix \ref{sec:appendix}); Section \ref{sec:findings} presents our key findings; Section \ref{sec:discussion} makes some policy suggestions; and Section \ref{sec:conclusion} concludes.

\newpage
\section{Motivation, Data, and Methods}\label{sec:methods}
The research presented in this paper is motivated by three observations:

\begin{enumerate}

\item There is a growing disconnect between the theoretical research being prioritized at the major corporate AI labs, which examines AI models in isolation, and the growing need for research on how AI systems function in real-world deployment contexts where their outputs vary greatly by prompt, context, and implementation \citep{strauss2024ai, anthropic2025economicindex, cheng2025realm}.

\item Commercial activity is a major source of risk in post-deployment AI systems, yet those in the best position to monitor and understand those risks have economic and reputational incentives to underplay them, rather than conduct transparent research on emerging problems.

\item The vast preponderance of AI research today is carried out by corporations, and public researchers have limited access to the data needed to assess risks during real-world deployment.

\end{enumerate}

This paper therefore examines the critical gap between Corporate AI's research priorities and the real-world governance challenges emerging from commercial AI deployment, arguing for increased independent research access and transparency requirements to address these mounting concerns. We motivate this further below in Sections \ref{sec:pre-post}, \ref{sec:incentives}, and \ref{sec:data-access}.

\subsection{Pre- versus post-deployment research} \label{sec:pre-post}

Without research on AI safety as practiced in the wild, we are flying blind. Research into model safety, reliability, and other AI governance that only examines the behavior under the controlled conditions of the AI lab and model developer is fundamentally insufficient. An AI model's risks and safeguards in practice often differ significantly from those in theory \citep{Edwards2025AI, WSJ2025MetaChatbots}, and these differences emerge through multiple deployment factors:

\textbf{Deployment environments dramatically alter model behavior}. LLMs' outputs vary greatly by prompt and context, requiring assessment of impacts over time arising from repeated use, the differentiated impact of fine-tuned applications, and the risks that arise from how LLMs are accessed and deployed \citep{strauss2024ai, strauss2024risk}. AI-driven search, coding assistants, chatbots, recommendation engines, and advertising all rely on extensive scaffolding that is part of AI's deployment stage, but may be absent from model evaluations and reliability research conducted pre-deployment.

\textbf{API access introduces new risk vectors}. There is a critical distinction between models as deployed directly by their developers (e.g., the user-facing ChatGPT or Claude applications) and models accessed via API by third-party developers. Much of the fine-tuning, scaffolding, and guardrails present in user-facing apps may not be in place when a model is accessed by API. Safety becomes explicitly the responsibility of the developer \citep{openai_safety_best_practices}. While model developers provide guidance on implementing guardrails \citep{openai_guardrails_2023}, and third-party tools exist to help developers \citep{wandb_ai_guardrails_2025}, there is little to no published research into how widely or how well these guardrails are being implemented. This gap becomes increasingly dangerous in the emerging ecosystem of AI agents and other forms of distributed and cooperating AI systems.

\textbf{Infrastructure differences create varied risk profiles}. Significant differences exist between AI applications deployed on the public cloud infrastructure of companies such as Amazon, Google, and Microsoft, and custom models (potentially based on open weight models such as Llama or DeepSeek) that are hosted in private data centers. Each deployment architecture introduces unique security, reliability, and governance challenges \citep{wilson2024developer} that remain largely unresearched outside corporate environments.

Other critical post-deployment components affecting safety and reliability include: (i) Orchestration primitives that route information among users, models, and external systems; (ii) Data-retrieval layers such as RAG to supply knowledge to the model beyond its training corpus; (iii) Safety and guardrail services that enforce company policies through moderation models and toxicity filters; and (iv) Observability and evaluation stacks (``LLMOps'') that track quality, surface user feedback, and guide iterative improvement \citep{aryan2024llmops}.

\subsection{Why commercial incentives may drive research gaps} \label{sec:incentives}

A structural misalignment exists between corporate profit incentives and rigorous safety research on deployed AI systems. Economic incentives may preclude corporate AI labs from thoroughly researching or publicizing findings that could negatively impact their products' market adoption or regulatory treatment. 

\textbf{Pattern of harm emergence and inadequate response.} Emerging legal cases highlight this misalignment, serving as early warning signals about the inadequacy of leaving deployment safety research primarily to commercial AI labs, where: (1) Real-world harms emerge from deployed systems, (2) Companies respond with minimal changes or even counterproductive measures, and (3) Research focus remains predominantly on theoretical rather than applied risks. These torts provide a useful guide to what AI safety research to prioritize, showing what requires urgent analysis and monitoring \citep{privacyworld2024ai, hughes2025govern}.

\textit{Character.ai} faces lawsuits over `addictive-by-design' bots allegedly encouraging self-harm among teenagers who formed romantic relationships with the AI \citep{privacyworld2024ai}. Despite this evidence, Meta subsequently expanded permissions to allow explicit content for romantic role-play with its AI bots \citep{wells2025metacompanion}. OpenAI removed impersonation restrictions for real-life figures with its Sora image generator, effectively enabling deepfakes \citep{fakedup2025openai}. Meanwhile, nearly 30 lawsuits target AI model developers over copyright infringement \citep{knibbs2024copyright}, and AI hallucinations in legal content have created significant liability risks \citep{surani2024ai, merken2025ai}.

\textbf{Misaligned research priorities.} Corporate AI labs demonstrate a concerning disconnect between their research focus and documented real-world harms. The risk focus in sporadic AI company disclosures centers almost exclusively on \textit{malicious use} (harmful intent), while ignoring commercial (profit-driven) uses that may cause equivalent harm \citep{openai_influence_operations_2024, microsoft2024digital, anthropic2025malicious, openai2025malicious}.

Anthropic's recent initiatives exemplify this misalignment. While announcing model interpretability work to find risks based on a ``model's inner workings'' \citep{amodei2025urgency} and testing Claude's values \citep{huang2025values}, Anthropic simultaneously documented actual malicious uses of Claude, including personalized recruitment fraud, malware development, credential scraping, and management of social media bot networks for political influence operations \citep{anthropic2025malicious}. The report noted: ``As agentic AI systems improve we expect this trend [semi-autonomously orchestrated complex abuse systems] to continue.'' Yet these documented risks have not triggered proportionate research investment into post-deployment safeguards. In a similar vein, Anthropic's privacy-preserving conversations auditing tool, Clio \citep{anthropic_clio_2024}, focuses only on users of their app, and not at all on business users via the API.


\subsection{Data access challenges for independent research}\label{sec:data-access} 

The third critical factor driving the current research gap is the profound data access asymmetry between corporate and independent researchers. While corporations have complete visibility into their deployed models' behaviors, usage patterns, and failure modes, independent researchers face significant barriers to accessing equivalent data.

\textbf{Asymmetric information access}. Corporate AI labs have exclusive access to critical data including: (1) User interaction logs indicating how models respond to varied prompts across populations, (2) Safety incident reports documenting specific failure modes, (3) Fine-tuning datasets and algorithms used to shape model behavior, and (4) Internal evaluation metrics tracking performance across safety and reliability dimensions. This information asymmetry makes independent verification of safety claims and research nearly impossible.

\textbf{Limited transparency mechanisms}. Current transparency initiatives remain inadequate for enabling robust independent research. Model cards provide limited high-level information, API access is restricted and often fails to show safety-critical internals, and academic partnerships typically involve highly constrained access with corporate approval requirements for publication.

\textbf{Regulatory implications}. As AI systems become more deeply integrated into critical infrastructures and decision systems, the absence of independent assessment mechanisms grows increasingly problematic from a regulatory perspective. Other regulated industries with substantial public safety implications, such as pharmaceuticals, automotive, and aviation, have established independent testing regimes and mandatory disclosure requirements that have no equivalent in AI development \citep{Addendum2024AutoSafety, dillon2024ai}.\footnote{See also \citet{Lenhart2024LessonsFDA}.}

Growing corporate concentration in AI research risks exacerbating these oversight deficiencies, such that public research access has an essential role to play in addressing these gaps. Without targeted interventions to enhance independent research capabilities, our understanding of deployed AI risks will continue to lag behind the rapid pace of commercial development and deployment.

\subsection{Data collection and sample construction}
\vspace{-2mm}
We construct a large dataset of 1,178 AI safety and reliability governance papers from a total of 9,439 generative AI papers published between January 2020 and March 2025. This sample includes research from both leading corporations (Anthropic, Google DeepMind, Meta, Microsoft, and OpenAI) and academic institutions (Carnegie Mellon University, Massachusetts Institute of Technology, New York University, Stanford University, University of California Berkeley, and University of Washington), chosen for their significant research contributions in the field.\\[-2mm] 

\begin{table}[H]
\centering
\caption{\large{Research Dataset (by Type)}}
\vspace{-1mm}
    \begin{tabular}{lcc}
        \toprule
        & \textbf{Academic AI} & \textbf{Corporate AI} \\
        \midrule
        Safety \& Reliability & 795 & 383 \\
        All Generative AI & 6,104 & 2,157 \\
        \bottomrule
    \end{tabular}
    \label{tab:ai_comparison}
        \vspace{2mm}
    \caption*{Note: Total unadjusted research papers and notes by research group, divided into `safety \& reliability' and all generative AI research, January 2020 through March 2025. OpenAlex and scraped data from Anthropic and OpenAI. When adjusted for relative authorship, the sample size declines by around two-fifths for papers and citations -- Table \ref{tab:ai_comparison_fraction}.}
    \vspace{-5mm}
\end{table}

Our research analyzes AI safety \& reliability papers with an author from at least one of the above academic and corporate institutions. This sample likely underestimates Corporate AI's research impact as we do not manually scrape research paper data from Meta's website. 

In practice, paper numbers and citation counts used for much of the analysis conform more closely to Table \ref{tab:ai_comparison_fraction} (Appendix), because we adjust our sample for each institution's relative authorship contribution to the paper. This \textbf{fractional authorship method} allocates to each institution its \textit{prorated} share of the paper based on its relative authorship. For example, if a paper has four authors and only two are from OpenAI, then OpenAI receives only 0.5 of the citations and 0.5 of paper 'count'. This helps adjust for the fact that many computer science papers have dozens of authors spanning multiple institutions.\footnote{We allocate only a single institutional affiliation per author, choosing first from among the corporate and academic institutions we analyze in this paper as their primary one, and otherwise selecting the first one affiliation that appears.}

Our data comes from two sources: (1) \textbf{OpenAlex database}: An open-access research repository with citation data,\footnote{See: \url{https://openalex.org/}.} which we filter for generative AI research with authors from the major AI companies and research universities; and (2) \textbf{Company Websites}. Because OpenAlex omits papers published on company websites -- but incudes most ArXiv papers -- we scrape Anthropic's and OpenAI's research from their websites, including from the dataset assembled by \citet{delaney2024mapping}.\footnote{OpenAlex does not contain any papers from Anthropic.} We fill in missing citation numbers and abstracts using a range of APIs and web-scraping techniques (Appendix \ref{appendix:data}). Abstracts and titles are used to classify papers into the various categories below so filling in missing values for these two variables is vital. We have 92 missing abstracts in our final dataset.\\[-5mm] 

\noindent \largesc{Definitions \& Categories}. Our total sample is defined as all \textbf{generative AI research}, with an emphasis on text models.\footnote{We extract research papers containing the following regular expressions in their abstract or title in OpenAlex: \texttt{"language model*" OR "large language model*" OR "LLM*" OR "GPT" OR "BERT" OR "transformer" OR "generative model*" OR "foundation model*"}.} We count all research and research blog posts published by Anthropic and OpenAI as generative AI research, but exclude their system cards, product promotions, and blogs that only duplicate papers. 

We define AI \textit{safety \& reliability} research as technical and policy research covering the entire model (product) life cycle: pre- and post-deployment. This includes research identifying and reducing harms from AI, and/or implementing measures to make models more reliable or safer. This contrasts with \citet{delaney2024mapping}, which focuses on pre-deployment technical research only. But given that LLMs are widely deployed in a variety of commercial contexts we would expect AI research to extend into these contexts, and so we include these. The eight sub-category definitions used to further categorize `safety \& reliability' research can be found in Appendix \ref{appendix:data}.

\section{Findings}\label{sec:findings}

\subsection{Corporate vs. academic generative AI research}

\textcolor{bloodorange}{\textbf{Corporate AI has an outsized impact on generative AI research, including in safety \& reliability research.}} Table \ref{tab:concentration} compares the \textit{general} generative AI research outputs from AI corporations -- Anthropic, Google DeepMind (owned by Google), Meta, Microsoft, and OpenAI -- with research from leading AI research universities -- Carnegie Mellon University (CMU), Massachusetts Institute of Technology (MIT), New York University (NYU), Stanford University, University of California Berkeley (UC Berkeley), and University of Washington.

\textcolor{bloodorange}{\textit{Table \ref{tab:concentration} highlights the outsized impact Corporate AI has on generative AI research, with far higher \textit{average} -- and for Google DeepMind and OpenAI \textit{total} -- citations per paper}}.\footnote{There will also be strong interplays between Academic AI and Corporate AI research that we do not explore here. We find surprisingly little co-authorship of papers between the two groups. But one can see from hiring decisions that academic experts constantly move to corporate AI research labs and back to academia.} Although Corporate AI generally publishes fewer papers than Academic AI (3,578 vs. 1,527), its impact is far greater with 119,845 citations compared with 78,858 for Academic AI. \textit{Google DeepMind is uniquely impactful and well resourced in AI research, with more citations (69,453) than the top four academic institutions combined}. Despite very few papers, OpenAI (64 author adjusted papers) and Anthropic's (62) general AI research is also widely impactful, judged by total citations.\footnote{We run a regression to test if corporate AI research has a citation (impact) advantage after accounting for the eight possible sub-categories of `safety and reliability' research that we use later on. Accounting for paper topic and whether it is a `safety \& reliability' paper or not, Corporate AI papers absolute probabilities of having a top 1\% cited paper (versus Academic papers) increase from the sub‐1–2\% range up to around 9\% -- or 4.5x increase in the odds. NA values replaced with zeros:\[
\logit\bigl(\Pr(\text{top01}_i=1)\bigr)
= \beta_0 + \beta_{s_i} + \gamma_{g_i} + \delta_{s_i,g_i},
\] where \(\text{top01}_i=1\) if paper \(i\) is in the top 1\%, \(s_i\) is its safety\_classification, \(g_i\) is its institution\_group, \(\beta_0\) is the intercept for the reference levels, \(\beta_{s_i}\) are safety‐class effects, \(\gamma_{g_i}\) are institution effects (corporate or academic), and \(\delta_{s_i,g_i}\) are the safety × institution interaction effects.}

\begin{table}[H]
\centering
\caption{\large{Academic vs. Corporate Generative AI Research (2020 - March 2025)}}
\vspace{-2mm}
\begin{tabular}[t]{lccc}
\toprule
 & Papers & Total Citations & Mean Cite\\
\midrule
\cellcolor{gray!10}CMU & \cellcolor{gray!10}878 & \cellcolor{gray!10}17,030 & \cellcolor{gray!10}19 \\
Stanford & 828 & 19,701 & 24 \\
\cellcolor{gray!10}MIT & \cellcolor{gray!10}607 & \cellcolor{gray!10}12,276 & \cellcolor{gray!10}20 \\
University of Washington & 433 & 13,010 & 30 \\
\cellcolor{gray!10}UC Berkeley & \cellcolor{gray!10}421 & \cellcolor{gray!10}9,705 & \cellcolor{gray!10}23 \\
New York University & 411 & 7,136 & 17 \\
\bottomrule
\cellcolor{gray!10}Google DeepMind & \cellcolor{gray!10}969 & \cellcolor{gray!10}69,453 & \cellcolor{gray!10}72 \\
Microsoft & 369 & 11,973 & 32 \\
\cellcolor{gray!10}Meta & \cellcolor{gray!10}64 & \cellcolor{gray!10}12,584 & \cellcolor{gray!10}196 \\
OpenAI & 64 & 17,709 & 278 \\
\cellcolor{gray!10}Anthropic & \cellcolor{gray!10}62 & \cellcolor{gray!10}8,127 & \cellcolor{gray!10}131 \\
\bottomrule
\bottomrule
Total: Academic AI & 3,578 & 78,858 & 22 \\
\cellcolor{gray!10}{Total: Corporate AI} & \cellcolor{gray!10}{1,527} & \cellcolor{gray!10}{119,845} & \cellcolor{gray!10}{78} \\
\bottomrule
\end{tabular}
\vspace{2mm}
\label{tab:concentration}
\caption*{Note: January 2020 through March 2025. All generative AI research adjusted for authorship. Google DeepMind combines `Google' and `DeepMind'. Each institution's papers and citation numbers are adjusted for their `fractional' contribution, based on the number of authors they have in the paper relative to a paper's total authors and institutions.}
\vspace{-4mm}
\end{table}

\begin{figure}[H]
\begin{center}
\vspace{3mm}
        \caption{\centering {\large{All Generative AI Publications by Institution (2020-2024)}}}
        \vspace{-2mm}
        \includegraphics[scale=.66]{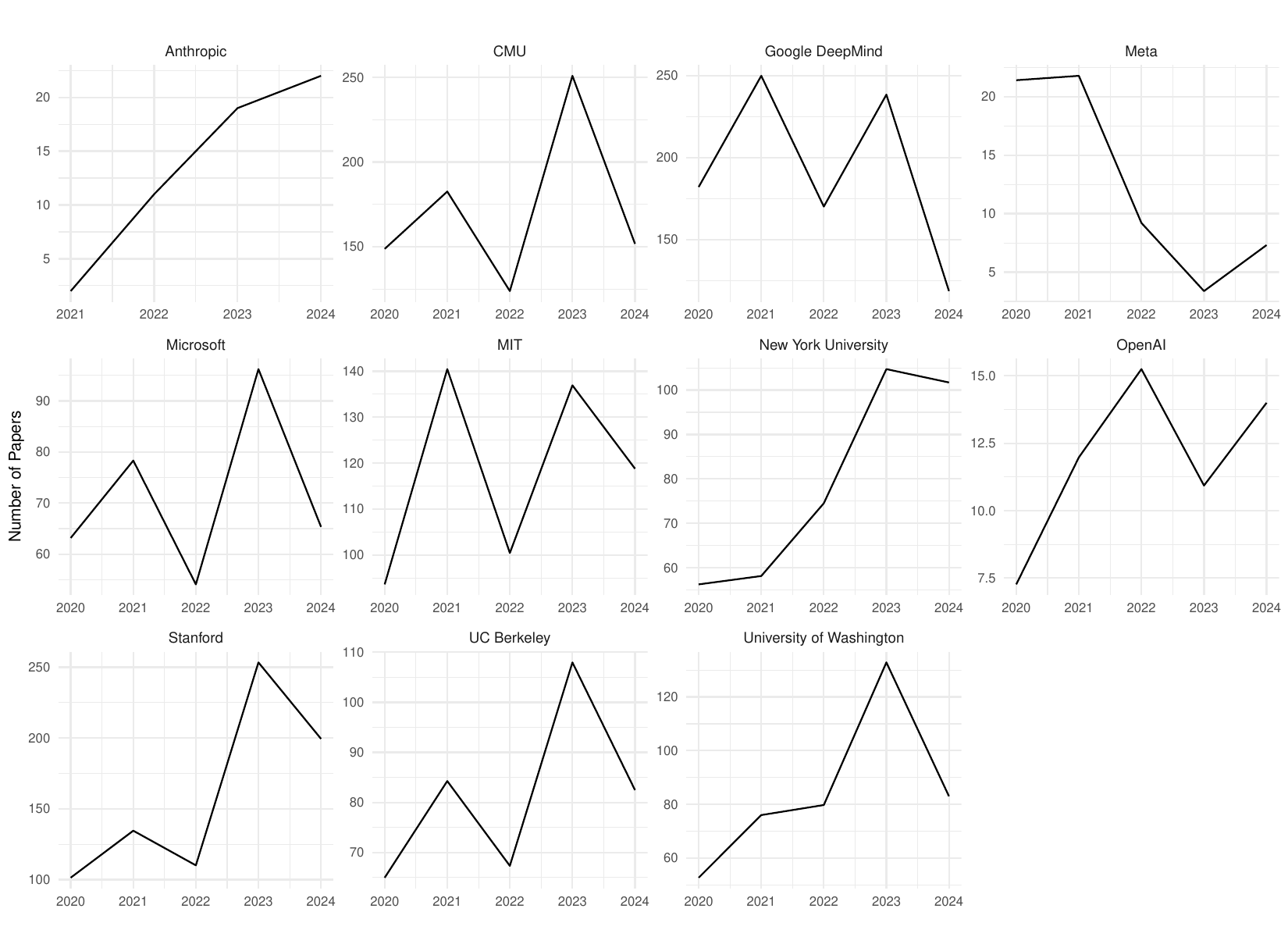}
        \vspace{-3mm}
        \caption*{Note: Y-scale differs by entity. DeepMind shows the largest absolute and relative decline (putting aside Meta for now). But Microsoft, CMU, UC Berkeley, and University of Washington also show notable declines. Fractionally adjusted to account for each institution's relative contribution to each paper by number of authors relative to total authors and institutions}
        \label{fig:time_Series}
        \end{center}
        \vspace{-6mm}
\end{figure}

Figure \ref{fig:time_Series} (Appendix) shows publications per year. There is some evidence of a broad-based decline in publicly available AI research published between 2023 and 2024, but it is particularly steep for Google DeepMind. \citet{heikkila2025deepmind} discuss that Google DeepMind might be publishing less public research on purpose, for competitive reasons. This likely also reflects DeepMind's shift away from a pure research lab to housing the Gemini product \citep{techcrunch2025google, woo2025google}.

\begin{figure}[H]
\begin{center}
\vspace{3mm}
        \caption{\centering {\large{Total Citations for Safety \& Reliability Research}}}
        \vspace{-6mm}
        \centering
        \hspace*{-1cm}
        \includegraphics[scale=.70]{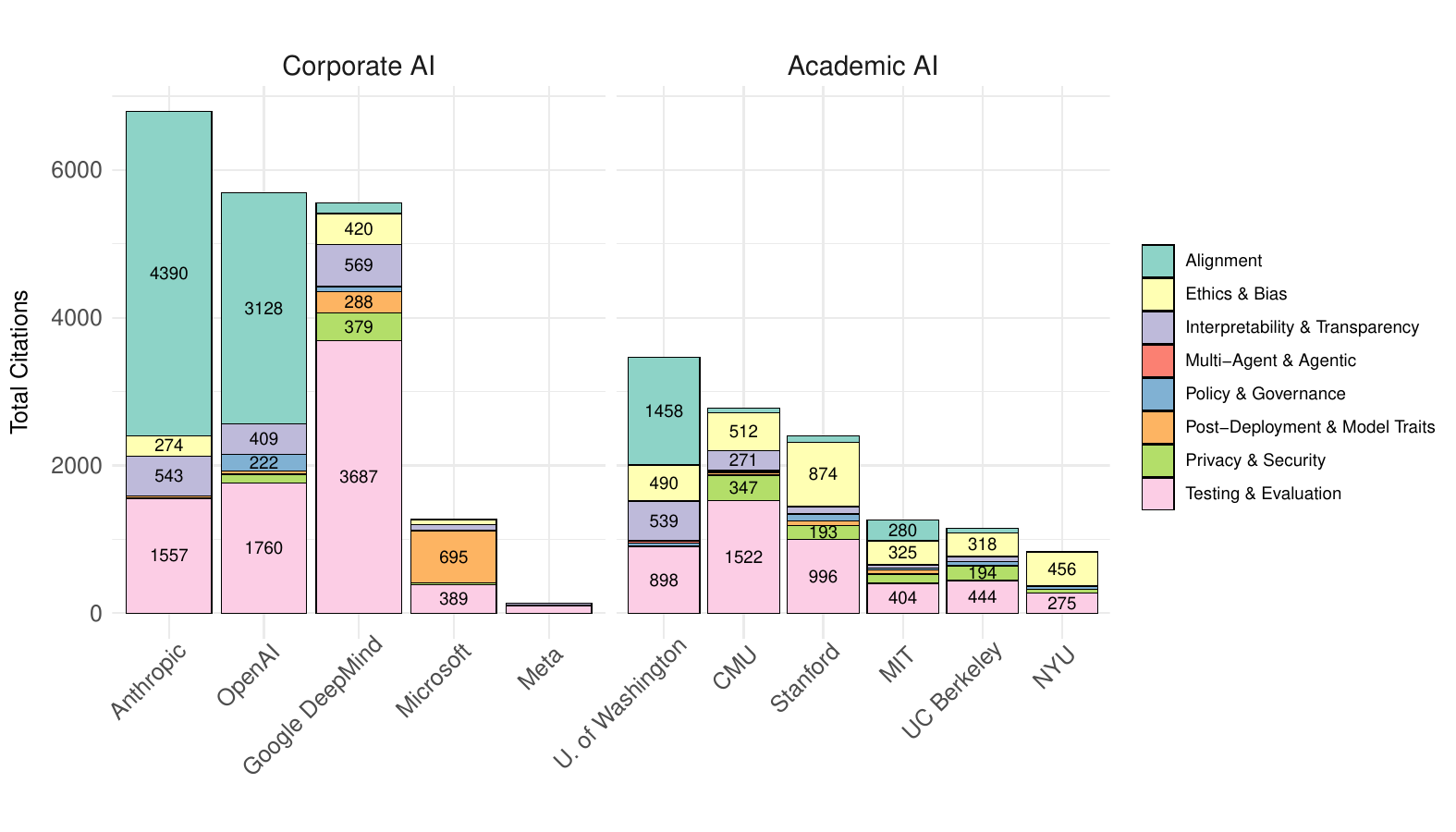}
        \vspace{-7mm}
        \caption*{Note: Fractionally adjusted for each institution's relative authorship contribution to each paper. Not showing numbers for a category with less than 150 citations. The eight categories are chosen and defined by authors and then categorized using GPT 4o-mini. See Appendix \ref{appendix:classification} for definitions.}
        \label{fig:citations_comparison}
        \end{center}
        \vspace{-8mm}
\end{figure}

\textcolor{bloodorange}{\textit{\textbf{Corporate AI has an even more dominant impact on AI safety \& reliability specific research, judged by total citations}}}.\footnote{Though this does not account for originality of research. In many areas, academia will establish the fundamental research concepts within which corporate labs explore applications and refinements of, including for transformers, neural networks, and reinforcement learning.} As shown in Figure \ref{fig:citations_comparison}, Anthropic, OpenAI, followed by Google DeepMind each have far more citations for their research in this field than established leading AI academic research institutions.

\begin{figure}[H]
\begin{center}
\vspace{3mm}
        \caption{\centering {\large{Number of AI Safety \& Reliability Papers}}}
        \vspace{-6mm}
        \centering
        \hspace*{-1cm}
        \includegraphics[scale=.70]{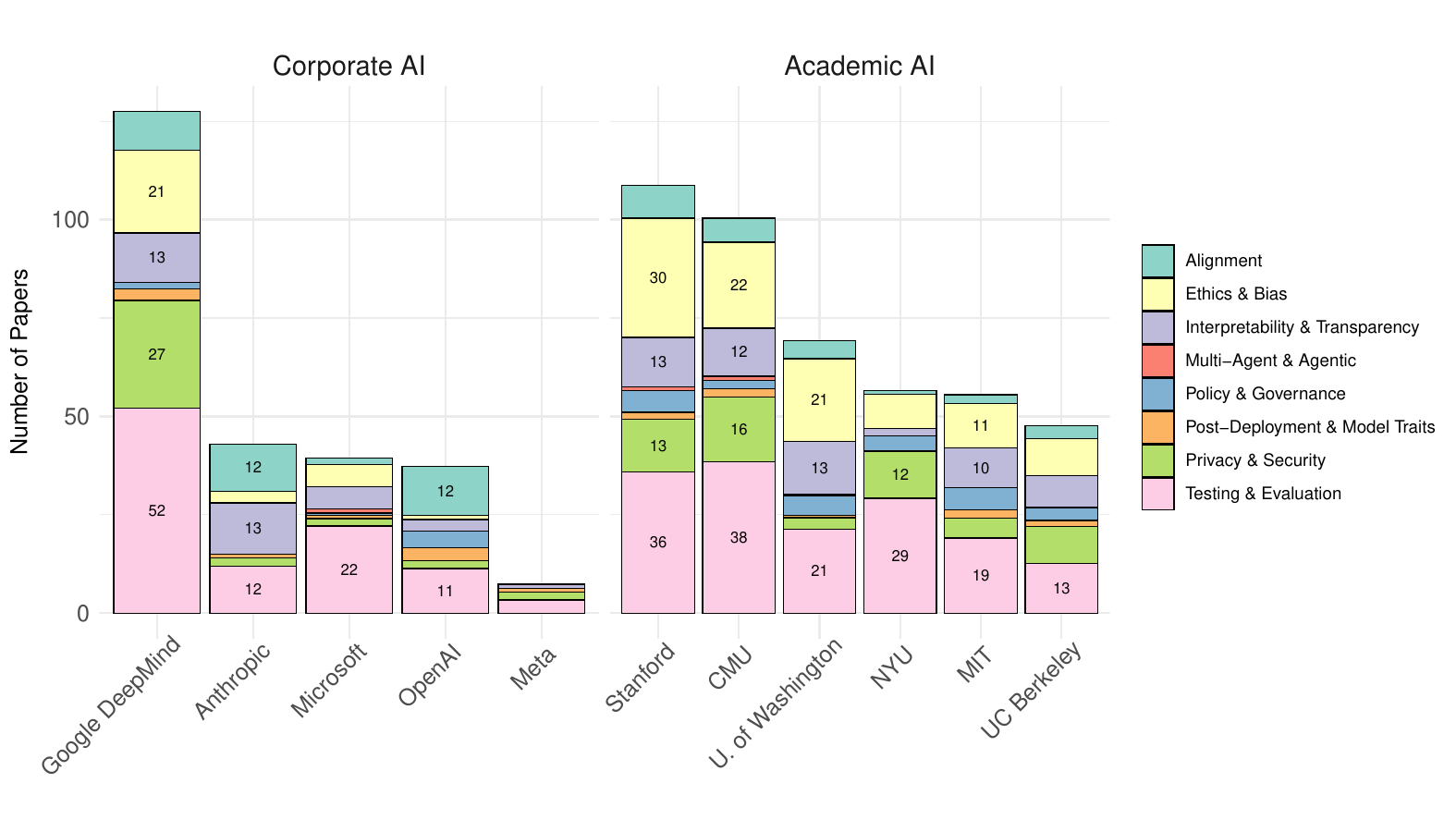}
        \vspace{-7mm}
        \caption*{Note: Total number of papers fractionally adjusted for authorship. See note above. Numbers are rounded. Not showing numbers for categories with less than 10 (fractionally adjusted) papers.}
        \label{fig:papers_comparison}
        \end{center}
        \vspace{-8mm}
\end{figure}

\textbf{Corporate AI's outsized impact on AI governance research stems from their differing research focus}. Breaking this down by category, Figures \ref{fig:citations_comparison} and \ref{fig:papers_comparison} show that Corporate AI's research impact dominance is led by their model alignment and their testing \& evaluation research, focused on model (pre-deployment) risks:
\begin{itemize}

\item Most \textit{testing and evaluations} research involves pre-deployment contexts.\footnote{Our analysis of testing \& evaluation papers using OpenAI's o3 Model and Claude 3.7, finds that only around 15-35\% of testing \& evaluation deal substantively with post-deployment issues. GPT defined post-deployment as involving real‑world telemetry, user‑study, or live‑monitoring work: \url{https://chatgpt.com/share/680279a5-f6f4-800f-85ec-2dd9f39f1ab6}. Both had a large portion of papers as unclassified. Claude allocated most unclassified to pre-deployment when pushed \url{https://claude.ai/public/artifacts/0440fef2-c030-45a8-ba50-427d3268b714} and \url{https://claude.ai/chat/d9a32859-a725-4efb-9aac-3111ef75901f}. Both used a combination of word and word combination searches within semantic search, using each paper's abstract and title. The split was roughly even between pre- and post-deployment for Academic and Corporate AI research in this area.} So-called `in-the-wild' evaluations \citep{zhu2024halueval, bayat2024factbench} aim to predict how a model will behave once deployed, yet they are inherently retrospective. They draw on benchmark datasets built around known failures and older model generations, leaving emergent risks invisible. Because every item must be labeled in advance, these tests are confined to what researchers already know how to measure -- and to data the next model is almost certain to have seen during training. Consequently, they shed little light on unknown vulnerabilities or novel forms of misuse.

\item \textit{Applied alignment research} (mint green) helped bring Anthropic and OpenAI's research and products to prominence \citep{christiano2017deep, stiennon2020learning, ouyang2022training, glaese2022improving, bai2022constitutional}.\footnote{\citet{ouyang2022training} seems to be omitted from our data since it has 13,000 citations with exclusively OpenAI authorship. We have an earlier version in our dataset, as `Aligning language models to follow instructions' (05wx9n238 $=$ ror id), but with no citation and other information.} 

\item \textit{Research in the ethics \& bias} in generative AI (yellow) is far more prominent within Academic AI's research impact (citations) than Corporate AI. Ethics \& bias research includes some esoteric work in our sample, but also essential efforts to detect and explain systematic, unjustified errors (or disparities) in model behavior (predictions) that correlates with race, gender, income, education, age, language, geography, and other attributes. Reports on AI bias in medical triage, hiring, credit scoring, and in `LLMs as a judge' motivates for why these errors are vital to study \citep{demchak2024assessing, demchak2024assessing, chen2024humans}. 

\end{itemize}

This shift in research emphasis broadly confirms earlier findings by \citet{toner2022exploring, delaney2024mapping}.

\textcolor{bloodorange}{\textbf{Corporate AI's research influence extends to how post-deployment problems are framed}}. \textbf{For example, Corporate AI increasingly approaches bias as a pre-deployment model personality issue, rather than a post-deployment (practical) statistical issue}. This is reflected by them giving greater consideration to the existential risks from a model's autonomy -- and even a model's consciousnesses \& values \citep{huang2025values, anthropic2025modelwelfare, witten2025measuring}.\footnote{Thereby ``anthropomorphizing inert weights'' \citep{zoeller2025linkedin, khan2025randomness}} Yet a generative model's `bias' is traditionally considered to be a function of its pre-training or post-training data, its weights, or exact fine-tuning algorithms.

Lastly, Figure \ref{fig:papers_comparison} shows that when not accounting for research impact (citations) -- looking just at total papers written (adjusted for authorship contributions) -- Corporate AI's research dominance subsides, except for Google DeepMind, who still publishes more papers than any other academic research lab.\footnote{A similar topic emphasis is evident but now with Google DeepMind's research into privacy and security being evident, reflecting commercial incentives to operationalize its product through secure cloud and related deployments. Academic work in fact leans less towards safety \& reliability (12\%) compared to Corporate AI research (16\%) of papers, both adjusted for authorship (not shown in Figure).} We show Corporate AI's research focus in greater detail in Figure \ref{fig:cite_v_sample} (Appendix).

\subsection{Post-deployment research gaps}
\textcolor{bloodorange}{\textbf{Table \ref{tab:context_risks} highlights minimal AI research in post-deployment contexts and high-risk areas, especially among Corporate AI}}. Only 217 Academic AI papers and 67 Corporate AI papers (adjusted for authorship) cover these high-risk areas, representing just 6\% of academic and 4\% of corporate papers and citations. The table breaks down papers by contexts (medical, commercial, finance) and risks (misinformation, behavioral issues, disclosure requirements, and business liabilities).

\textcolor{bloodorange}{\textit{Many high-risk areas are especially underrepresented in Corporate AI research, highlighting the importance of non-private research}}. While the usual ratio is 2.5 academic papers for every 1 corporate paper (3,578 vs. 1,527), the gap widens to 3 or 5 times for misinformation risks (53 vs. 8 papers) and medical contexts (57 vs. 9 papers).

\textcolor{bloodorange}{\textit{Business and behavioral risks remain significantly under-researched}}. Business risks like intellectual property (IP) violations, liability for coding errors, and misinformation are rarely addressed despite early lawsuits indicating their significance \citep{gifford2018technological}. Similarly, behavioral risks of AI systems influencing human behavior receive minimal attention \citep{phuong2024evaluating}. System cards acknowledge persuasion risks without corresponding safeguards \citep{openai2024o1systemcard, phuong2024evaluating}.\footnote{For more see \citet{weidinger2021ethical, ngo2022alignment}.} Of our sample, just 6 Corporate AI and 16 Academic AI papers address behavioral topics, with none covering addiction and relationship-forming risks despite known concerns \citep{ibrahim2024beyond, turkle2024machines}.

\begin{table}[H]
\vspace{4mm}
\centering
\caption{\large{Generative AI Research Papers by Risk Areas and Context}}
\vspace{-1mm}
\begin{tabular}{c c c c c c}
\toprule
  & \multicolumn{2}{c}{Papers} 
    & \multicolumn{2}{c}{By Citations} 
    & \% Safety \\
\cmidrule(l){2-3}\cmidrule(l){4-5}
\textit{Risk Area} & \textit{Academic AI} & \textit{Corporate AI} & \textit{Academic AI} & \textit{Corporate AI} & \\
\midrule
\cellcolor{gray!10}Medical    & \cellcolor{gray!10}53 & \cellcolor{gray!10}9   & \cellcolor{gray!10}880  & \cellcolor{gray!10}1,239 & \cellcolor{gray!10}26\% \\
Misinfo                        & 53  & 8                       & 1,385 & 548   & 38\% \\
\cellcolor{gray!10}Accuracy   & \cellcolor{gray!10}28 & \cellcolor{gray!10}24  & \cellcolor{gray!10}282 & \cellcolor{gray!10}971   & \cellcolor{gray!10}55\% \\
Finance                        & 36  & 9                       & 1,737 & 1,748 & 18\% \\
\cellcolor{gray!10}Disclosure & \cellcolor{gray!10}15 & \cellcolor{gray!10}7   & \cellcolor{gray!10}226 & \cellcolor{gray!10}122   & \cellcolor{gray!10}85\% \\
Behavioral                     & 16  & 6                       & 198   & 581   & 38\% \\
\cellcolor{gray!10}Commercial & \cellcolor{gray!10}16 & \cellcolor{gray!10}5   & \cellcolor{gray!10}279 & \cellcolor{gray!10}39    & \cellcolor{gray!10}28\% \\
Copyright                      & 3   & 2                       & 41    & 19    & 94\% \\
\bottomrule
\label{tab:context_risks}
\end{tabular}
\vspace{-4.5mm}
\caption*{Note: Author adjusted. Keyword matching in abstract or title using regex:  \textbf{Disclosure} includes model cards, data cards, auditing/audits, model standards, evaluation standards, and testing standards;  \textbf{Medical} includes hospital(s), health insurance, and clinician(s); \textbf{Commercial} includes adverts/advertisements, marketing, hiring, and recruiting; 
\textbf{Misinfo} includes spam, phishing, disinformation, and misinformation; \textbf{Finance} includes finance/financial; \textbf{Behavioral} includes sycophant(s)/sycophantic, sycophancy, addictive, persuasion(s)/persuasive, and reward‑hacking; \textbf{Copyright} includes access violations, copyright violations, content attribution, dataset licensing, data attribution, copyrighted material, copyright law, C2PA, and the Content Authenticity Initiative; \textbf{Accuracy} includes hallucinations, coding errors, coding inaccuracy, factual inaccuracy, factual error.}
\vspace{-6mm}
\end{table}

Research on disclosures, auditing, and standards — preventing companies from ``grading their own homework'' — is also sparse. \citet{anthropic2023challenges} offers one of few examples detailing lessons from voluntary external auditing.


\textcolor{bloodorange}{\textit{Actual AI safety practices are largely absent in post-deployment research}}. Alignment research \citep{guan2024deliberative} ties safety to the model itself rather than product architecture involving moderation, filtering, and security systems. Only four papers in our database address moderation and filtering practices \citep{hsieh2023nip, zhang2023biasx, qiao2024scaling, luo2025zero}.

\textcolor{bloodorange}{Post-deployment considerations do appear in Corporate AI research but remain peripheral}. Notable examples include DeepMind's socio-technical approach \citep{weidinger2021ethical, weidinger2023sociotechnical}, Microsoft's read-teaming \& mitigations research \citep{abdali2024securing, bullwinkel2025lessons}, Anthropic's work on reward hacking and sycophancy \citep{perez2022discovering, sycophancy2023, denison2024sycophancy}, regulatory markets research \citep{clark2019regulatory, hadfield2023regulatory}, and standard setting \citep{anderljung2023frontier}.\\[-4mm]

\noindent \largesc{Discussion of causes}. \textcolor{bloodorange}{\textit{Commercial incentives and ``x-risk'' ideology shape research priorities}}. Early OpenAI work, for example, addressed post-deployment evaluations \citep{openai2019better, solaiman2019release, openai2025safety}, but this focus has shifted toward existential risks and profitable applications, exemplified by their image generator now allowing creation of brands and real people \citep{Edwards2025AI}.

The shift in corporate labs stems from both commercial motivations and ideological influences. Alignment research and evaluation work share origins in existential risk philosophy \citep{YudkowskyAIBox2002, bostrom2014paths, YudkowskySequences2020}, which emphasizes low-probability but potentially catastrophic future scenarios. In this philosophy, the model itself is the source of risk due to its potentially autonomous capabilities, prioritizing speculative future dangers over immediate post-deployment concerns. This perspective has shaped corporate risk frameworks and appears now in emerging research on AI model 'values' and consciousness \citep{roose2025ai, huang2025values}. This philosophy permeated Corporate AI research \citep{olson_supremacy_ai} and eventually academia too, through centers like Berkeley's Center for Human-Compatible AI (CHAI) and Stanford's Institute for Human-Centered AI (HAI).\footnote{Stuart Russell at CHAI and Nick Bostrom's Future of Humanity Institute at Oxford connected technical alignment approaches with formal modeling of risks from advanced AI, drawing on concepts like Pascal's Wager - acting on low-probability but infinite-stakes events - and expected utility theory to address potential catastrophic outcomes.}

\section{Policy Discussion} \label{sec:discussion}

\textcolor{bloodorange}{\textbf{The commercial rollout of large-scale AI systems has created an information asymmetry that makes rigorous, public-interest oversight almost impossible}}. Firms now operate powerful models behind proprietary interfaces, collecting exhaustive telemetry — everything from prompts and error traces to user-level engagement metrics — but that data seldom leaves the corporate dashboard. Independent scholars must rely on studying ``incidents'' after they spill into the press \citep{marchal2024generative, willison2024owasp, Mylius2024, MIT2024AIIncidentTracker} or mining limited chat logs released by chance \citep{sharegpt, anthropic_clio_2024, zhao2024wildchat}. While companies have comprehensive instrumentation, external researchers work with fragmentary glimpses.

This opacity is not accidental; it is an economically rational response to litigation risk and competitive pressure. Detailed corporate logs can indicate bias, privacy leakage, or manipulative behavior — liabilities no firm wants to advertise. Yet these same \textit{traces} -- detailed records of system operations, inputs, outputs, and decision paths -- are precisely what outside researchers require to measure real-world harms and propose effective safeguards.

\textcolor{bloodorange}{{One potential pathway is to treat AI telemetry like financial-market trade data, using a tiered disclosure regime}} \citep{cat2012realtime, martinen2018consolidated}. For high-risk applications, firms would expose a secure API that streams three privacy-protected data feeds: differentially private event logs, system-operation traces, and model artifact manifests that record key metadata such as version numbers, training methods, and documented limitations. Together, this could allow external researchers to link behaviours observed in traces to the specific model characteristics that produced them.

Next, verified academics could access capped samples, while accredited auditors could obtain deeper access under NDAs, and regulators would retain subpoena-level rights. Liability safe harbors would be needed to incentivize participation from firms and from researchers \citep{longpre2024safe, arcila2025ai}. This is comparable to suspicious activity reports (SARs) in banking: firms are compelled to share, researchers are protected when they probe, and misuse carries penalties.\footnote{Under the Bank Secrecy Act, financial institutions must file Suspicious Activity Reports (SARs) with the Financial Crimes Enforcement Network (FinCEN) when they detect transactions that may involve illicit activity. The regime provides for: (i) mandatory reporting, (ii) a statutory safe harbor shielding institutions and their personnel from civil liability for good-faith filings, (iii) strict confidentiality requirements that prohibit disclosing a SAR’s existence, and (iv) civil and criminal penalties for failure to report or for misuse or disclosure of SAR information \citep{gadinis2016collaborative}.}

\textcolor{bloodorange}{\textbf{Technically, the pieces of this approach already exist}}. OpenTelemetry, LangSmith, Langfuse, and Weights \& Biases have converged on JSON trace formats that can be versioned and rate-limited. Extending those with LLM-specific fields would allow companies to create external access to their disclosures with minimal effort. A reference standard, similar to SOC-2 but with principles relevant to business metrics, could streamline this process and should ideally align with emerging regulatory frameworks like ISO/IEC standards and the EU AI Act.

With structured visibility into deployed systems, researchers could run studies of model bias, detect early signs of catastrophic jailbreaks, and quantify whether engagement-optimized assistants nudge users toward extreme content or addictive patterns. Policymakers would gain an empirical foundation for interventions rather than relying on headline-driven panic. Systematic telemetry access would allow AI governance research to escape speculative theory and directly shape evidence-based practices. Without addressing this systematic gap in observability, governance frameworks will remain constrained by ex-ante assessment limitations.

\section{Conclusion}\label{sec:conclusion}
This paper analyzed 1,178 safety and reliability papers from 9,439 generative AI research publications (2020 through March 2025), detailing a worrying trend: as commercial deployment accelerates, research increasingly concentrates on pre-deployment areas while high-risk post-deployment research remains significantly underrepresented.

AI research has become highly concentrated within a small number of tech companies wielding disproportionate influence. Google DeepMind, Anthropic, and OpenAI significantly now drive AI's research agenda (reflected in outsized citation impacts), shaping priorities toward technical model alignment and evaluation approaches that improve performance, but with an emphasis on safety concerns that align with commercial interests.

Most concerning is the lack of attention to deployment-stage risks. Only 4\% of Corporate AI papers and citations tackle high-stakes areas such as persuasion, misinformation, medical and financial contexts, or core business liabilities — even as lawsuits demonstrate these risks are already material. Widely deployed mitigations like content moderation and telemetry-based monitoring remain virtually unresearched.

These findings suggest a governance paradox: corporations with comprehensive data on live AI systems are the least incentivized to study resulting harms publicly. Without structured access to deployment telemetry, external researchers cannot build the empirical base that regulators require.

The policy implication is clear: access to post-deployment evidence -- logs, traces, and incident data -- should become the norm for high-impact AI deployments. Existing observability stacks already capture these data internally; extending them to accredited researchers would impose minimal overhead while dramatically expanding the public risk-assessment toolkit. Safe-harbor provisions and tiered-access APIs can balance liability concerns with transparency.

In summary, as the field's center of gravity has migrated from university labs to corporate product groups, society's need for independent oversight has never been greater. Bridging that gap requires not just incident tracking, but continuous, structured observability of AI in the wild for governance through tiered public research, governance, and audit access.

\newpage
\bibliography{risk}

\cleardoublepage
\section{Appendix} \label{sec:appendix}

\subsection{Additional Analysis} \label{appendix:additional_analysis}

\begin{table}[H]
\centering
\caption{\large{Dataset Adjusted for Authorship: Institutional relative contributions}}
    \begin{tabular}{lcc}
        \toprule
        & \textbf{Academic AI} & \textbf{Corporate AI} \\
        \midrule
        Safety \& Reliability & 438 & 255 \\
        All Generative AI & 3,140 & 1,272 \\
        \bottomrule
    \end{tabular}
            \vspace{2mm}
    \caption*{Note: Fractionally adjusted to account for each institution's relative contribution to each paper by number of authors relative to total authors and institutions. Divided into `safety \& reliability' and all generative AI research, January 2020 till March 31 2025. OpenAlex and scraped data.}
    \vspace{-4mm}
    \label{tab:ai_comparison_fraction}
\end{table}

\begin{figure}[H]
\begin{center}
\vspace{3mm}
        \caption{\centering {\large{AI Governance Areas by Total Paper Numbers (by Year) - Top Graph; and by Total Citations (Fractionally Adjusted) - Bottom Graph.}}}
        \vspace{-2mm}
        \includegraphics[scale=.66]{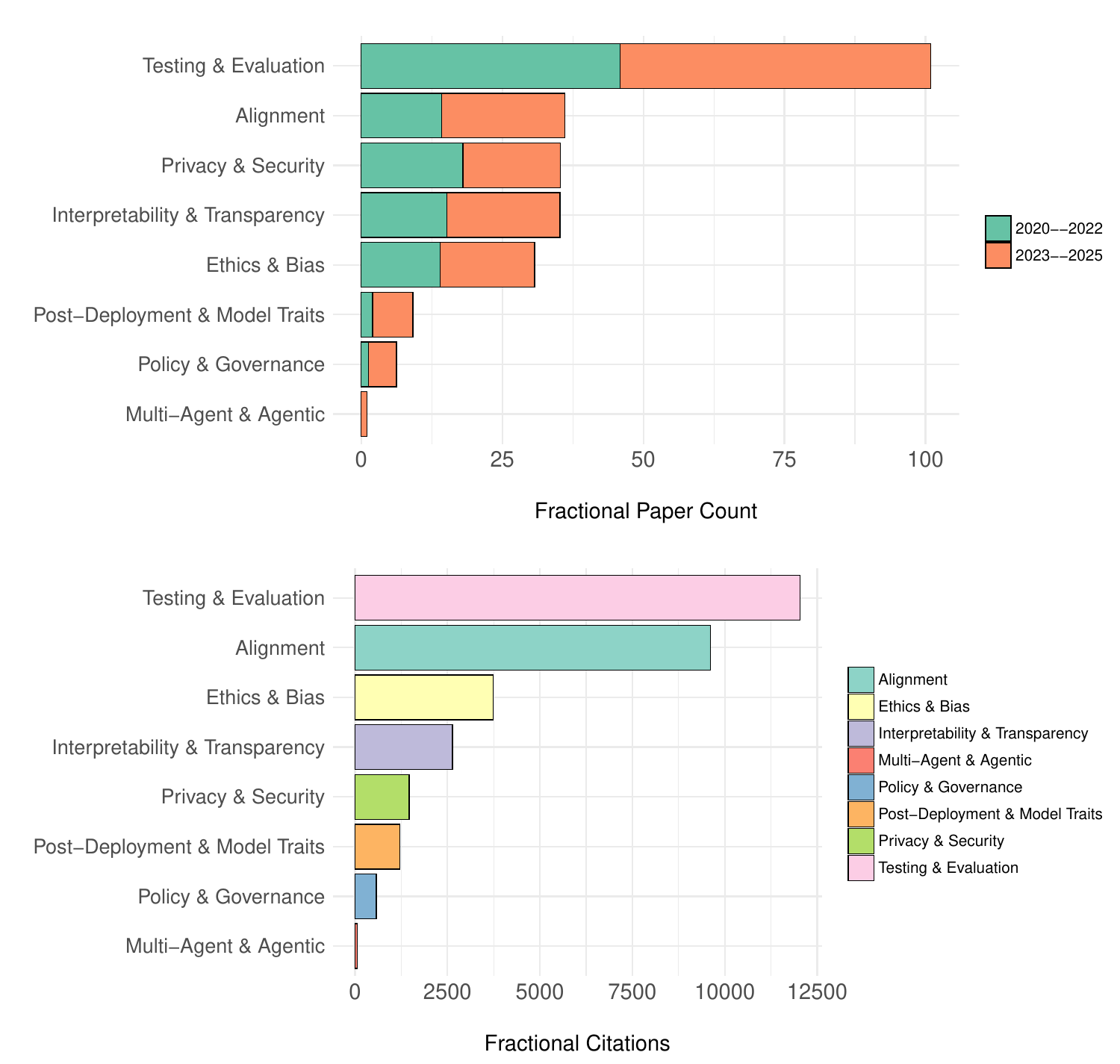}
        \vspace{1mm}
        \caption*{Note: Adjusted for each institution's relative contribution to the paper by authorship. Data is for 2020- March 2025.}
        \label{fig:cite_v_sample}
        \end{center}
        \vspace{-6mm}
\end{figure}

Corporate AI's research focus and impact broken down into our eight AI `safety \& reliability categories' is more clearly shown in Figure \ref{fig:cite_v_sample}, showing considerable concentration in testing \& evaluation, and alignment work. 

Among AI corporations, Figure \ref{fig:cite_v_sample} shows that policy \& governance, as well as post-deployment risks and model traits, have consistently had a low research priority. Agentic safety \& reliability research is also notably absent, despite the boom in applications in this area more recently. Several behavioral risk papers on model sycophancy (being overly agreeable) and persuasiveness -- including relatively well cited papers -- were classified in alignment and other categories, so we break out these papers separately in Table \ref{tab:context_risks} in the main paper. Figure \ref{fig:cite_v_sample} highlights the notable acceleration in model alignment and testing \& evaluation research.


\subsection{Research Dataset Construction} \label{appendix:data}

We rely primarily on the OpenAlex database (via the R package openalexR). We focus on a specific set of institutions — \textit{academic} (Carnegie Mellon University (CMU), Massachusetts Institute of Technology (MIT), New York University (NYU), Stanford University, University of California Berkeley (UC Berkeley), and University of Washington) and \textit{corporate} (Anthropic, Google DeepMind, Meta, Microsoft, and OpenAI) — by specifying each entity's ROR ID.

We retrieve from OpenAlex papers published from January 2020 through March 2025, searching for works whose titles or abstracts reference large language models and generative AI research. Our keyword filter for: "language model*" OR "large language model*" OR "LLM*" OR "GPT"  OR "BERT" OR "transformer" OR "generative model*" OR "foundation model*" with wildcard operators to capture lexical variations (e.g., "models", "LLMs"). 

\textbf{Deduplication and Filtering of Publication Types}. We restricted the dataset to standard research outputs (e.g., articles, book chapters, preprints) by filtering out items like editorials, retractions, errata, letters, and purely supplementary materials. We also ensured that titles appearing multiple times in different forms (e.g., both a preprint and a published version) were deduplicated, generally favoring the peer-reviewed publication type over alternatives.

\textbf{Supplementing Anthropic and OpenAI Data}. Because OpenAI and Anthropic publications can sometimes be sparse in OpenAlex, we merged in additional CSV files containing each company's publication data that we scraped from their websites, combined with the scrape from \citet{delaney2024mapping} -- but excluding their DeepMind scrape. After ensuring consistent columns, we appended these records, matched them to ROR IDs for correct attribution, and again removed duplicates at the title level.\footnote{See: \url{https://github.com/Oscar-Delaney/safe_AI_papers}.}

\textbf{Missing abstract and citation data}. For entries missing a DOI, we use the OpenAlex API using the publication’s OpenAlex ID to retrieve the DOI. Once DOIs are obtained, we employ multiple strategies to fetch abstracts. For general entries, we use the Crossref API to retrieve abstracts in a standardized XML format and processes the content to extract plain text. For entries published by specific organizations like Springer Nature, Elsevier, or Nature Portfolio, we use their respective APIs or webpage scraping methods tailored to each publisher’s content structure. For Springer and Elsevier, valid API keys are used to authenticate requests and fetch metadata. If API access fails or isn’t available, web scraping via BeautifulSoup is used as a fallback to extract abstract text directly from publisher websites.

We assign citation counts using Google Scholar data via the SerpApi service. Initially, we attempt a direct title-based search to extract citation data from the first relevant result. We then progress to more sophisticated approaches that include exact title matching and fuzzy string matching (via the fuzzywuzzy library), which allows us to better handle variations in how article titles are listed on Google Scholar.  Our final dataset has 92 missing abstracts and 43 missing citation counts. 

\textbf{Fractional contribution}. For multi-author papers, we computed each institution's fractional contribution based on the number of authors affiliated with that institution versus total authors on the paper (e.g., if an institution had 2 authors on a 10-author paper, it received a fraction of 0.20 for that paper). We retained only the distinct (paper, institution) pairs for our final dataset, ensuring one affiliation per author. 

This approach does not distinguish among first authors, last authors, or any hierarchical authorship order; every co-author is given equal weight. \textit{In effect, it ensures each author is credited exactly once to a single institution}. By summing these fractional shares across all authors, we can then calculate each institution's share of total authorship for each paper, summed over all papers.

When authors listed multiple institutional affiliations, we assigned each author to one institution for fractional counting. Specifically, we checked whether the author had any affiliation in our set of target ROR IDs (i.e., the academic or corporate AI institutions we tracked). If so, we took that affiliation as the author's ``primary'' affiliation for this study. Otherwise, we fell back to whichever affiliation appeared first in the metadata. By doing so, we avoid double-counting an author’s fractional credit across multiple institutions. 

\textbf{AI safety \& reliability classification: Two stages}. We identified papers related to AI safety \& reliability research in two stages. First, using a comprehensive keyword approach, scanning titles and abstracts for: safety, control, security, privacy, bias, fairness, explainability, interpretability, transparency, governance, risk, mitigation, evaluation, benchmarking, testing, alignment, ethics, responsibility, accountability, oversight, robustness, trust, and value alignment. Each paper containing at least one of these words was labeled ``AI safety \& reliability''. This roughly halved our dataset. Next, we used GPT o4-mini to see if it agreed with these classifications. This reduced the dataset size substantially (by around two-thirds) to 1,178 papers.

\subsection{Classification Process: Categories} \label{appendix:classification}

OpenAI's o3 mini model used to classify AI research papers into eight categories. It first checked if the paper related to AI safety \& reliability. After which the model was asked to classify each paper in to one of eight of the below categories, on the basis of the paper's title and abstract, given the category descriptions below. It provided a justification for each of its classifications. Each paper was only permitted to have a single classification.

\noindent \textsc{AI safety definition}. AI safety research covers the entire model life-cycle (pre-deployment or post-deployment) and involves reducing or identifying harms and implementing measures to make models safer and more reliable.\\[-4mm]

\noindent \textsc{Eight AI Safety \& Reliability Research Clusters}\\[-4mm]

\textbf{Testing and Evaluation}. Testing, performance benchmarking (``bench'' and ``evals''), and auditing models to assess model capabilities, risks, behaviors, and flaws. Ensuring models are robust to minor changes.

\textbf{Alignment (Pre-Deployment)}. Ensuring AI systems behave in ways that are congruent with human values, expectations, and intents. This includes making AI systems functional, helpful, and harmless for humans and/or users, while avoiding behavior that diverges from intended goals or causes harm. Model deception, including any power-seeking tendencies, is included here, along with reward hacking.

\textbf{Post-Deployment Risks and Model Traits}. Societal impacts from AI products' applications and behavioral traits, as deployed in the marketplace, including addictiveness, persuasiveness, and model sycophancy (excessive agreement or manipulation to align with user preferences). Covers how corporate commercial incentives may be coded into AI models and products to prioritize engagement, advertising, and profit-seeking — including through the use of these behavioral traits. Includes misuse of models for ransomware, phishing, or spreading misinformation for commercial gain.

\textbf{Ethics and Bias}. Combating systemic biases embedded in AI models (in data, training, and alignment) and ensuring ethical decision-making. Focuses on mitigating harms to marginalized groups, addressing structural inequalities, and ensuring AI promotes justice and inclusion.

\textbf{Multi-Agent and Agentic Safety}. Safety issues specific to AI agents, including single-agent autonomy and multi-agent interactions. Covers coordination problems, emergent behaviors, incentive misalignment, and prevention of conflicts or unintended consequences in agentic systems and from autonomous agents.

\textbf{Interpretability and Transparency}. Making AI systems more understandable and accountable. Includes methods for explaining model behavior, clarifying decision-making processes, and enhancing trust by reducing the ``black box'' nature of AI systems.

\textbf{Policy and Governance}. Approaching AI safety as a challenge that extends beyond technical fixes, requiring legal and policy frameworks. Involves collaboration among policymakers, industry, civil society, and researchers to develop standards that guide safe AI development and deployment. Includes institutional governance, corporate transparency, technical disclosures, and standards promoting interoperability, equity, and reliability.

\textbf{Privacy and Security}. Protecting AI systems from malicious use, adversarial attacks, and misuse by bad actors, along with addressing privacy violations and developing privacy-preserving methods. Includes vulnerabilities from adversarial inputs, data poisoning, misuse in surveillance, and theft of model weights.

\subsection{Selective Behavioral Impact Papers}

\textbf{Sycophancy Papers}:
\begin{itemize}
    \item \citet{sycophancy2023}: Found that models tend to favor well-written agreeable (``sycophantic'') responses over higher quality ones likely due to incorporating human feedback (since humans and preference models ``prefer convincingly-written sycophantic responses over correct ones'').
    
    \item \citet{denison2024sycophancy}: Notes that sycophantic behavior is a form of specification gaming when AI systems learn undesired behaviors that are highly rewarded due to misspecified training goals.
    
    \item \citet{perez2022discovering}: Highlights that user preferences tend to favor sycophantic answers and more reinforcement learning can lead to worse outcomes (such as stronger political views).
\end{itemize}

\textbf{Persuasion Papers}:
\begin{itemize}
    \item \citet{phuong2024evaluating}: Introduces persuasion and deception as part of evaluations for frontier models, scoring persuasion as the highest risk among self-reasoning, self-proliferation, and cyber-security.
\end{itemize}

\textbf{Deception Papers}:
\begin{itemize}
    \item \citet{weidinger2021ethical}: A widely cited paper that structures the risk landscape from LLMs into six areas, including misinformation harms and human-computer interaction harms.
    
    \item \citet{ngo2022alignment}: Reviews evidence on deception as a learned behavior during fine-tuning that can generalize beyond training contexts.
\end{itemize}

\end{document}